\pdfoutput=1
\documentclass[11pt,a4paper]{article}
\usepackage{times}
\usepackage{latexsym}
\usepackage{hyperref}
\usepackage{cleveref}
\usepackage[hyperref]{acl2017}
\usepackage{amsmath}
\usepackage{amsfonts}
\usepackage{graphicx}
\usepackage{url}
\aclfinalcopy 
\def\aclpaperid{3250} 

\title{A Conditional Variational Framework for Dialog Generation}
 \author{Xiaoyu Shen$^1$\thanks{Authors contributed equally. Correspondence should be sent to  H. Su (suhui15@mails.ucas.ac.cn) and X. Shen (xshen@lsv.uni-saarland.de).}, Hui Su$^2\footnotemark[1]$, Yanran Li$^3$, Wenjie Li$^3$, Shuzi Niu$^2$, Yang Zhao$^4$\\
{\bf  Akiko Aizawa$^4$ and Guoping Long$^2$} \\
$^1$Saarland University, Saarbr\"ucken, Germany\\
$^2$Institute of Software, Chinese Academy of Science, China\\
$^3$The Hong Kong Polytechnic University, Hong Kong\\
$^4$National Institute of Informatics, Tokyo, Japan\\}

\date{06/02/2017}

\begin{document}
\maketitle
\begin{abstract}
  Deep latent variable models have been shown to facilitate the response generation for open-domain dialog systems. However, these latent variables are highly randomized, leading to uncontrollable generated responses. In this paper, we propose a framework allowing conditional response generation based on specific attributes. These attributes can be either manually assigned or automatically detected. Moreover, the dialog states for both speakers are modeled separately in order to reflect personal features. We validate this framework on two different scenarios, where the attribute refers to genericness and sentiment states respectively. The experiment result testified the potential of our model, where meaningful responses can be generated in accordance with the specified attributes.

\end{abstract}

\section{Introduction}
Seq2seq neural networks, ever since the successful application in machine translation~\cite{sutskever2014sequence}, have demonstrated impressive results on dialog generation and spawned a great deal of variants~\cite{vinyals2015neural,yao2015attention,sordoni2015neural,shang2015neural}. The vanilla seq2seq models suffer from the problem of generating too many generic responses (generic denotes safe, commonplace responses like ``I don't know''). One major reason is that the element-wise prediction models stochastical variations only at the token level, seducing the system to gain immediate short rewards and neglect the long-term structure. To cope with this problem, \cite{serban2016hierarchical} proposed a variational hierarchical encoder-decoder model (VHRED) that brought the idea of variational auto-encoders (VAE)~\cite{kingma2013auto,rezende2014stochastic} into dialog generation. For each utterance, VHRED samples a latent variable as a holistic representation so that the generative process will learn to maintain a coherent global sentence structure. However, the latent variable is learned purely in an unsupervised way and can only be explained vaguely as higher level decisions like topic or sentiment. Though effective in generating utterances with more information content, it lacks the ability of explicitly controlling the generating process.

This paper presents a conditional variational framework for generating specific responses, inspired by the semi-supervised deep generative model~\cite{kingma2014semi}. The principle idea is to generate the next response based on the dialog context, a stochastical latent variable and an external label. Furthermore, the dialog context for both speakers is modeled separately because they have different talking styles, personality and sentiment. The whole network structure functions like a conditional VAE~\cite{sohn2015learning,yan2016attribute2image}. We test our framework on two scenarios. For the first scenario, the label serves as a signal to indicate whether the response is generic or not. By assigning different values to the label, either generic or non-generic responses can be generated. For the second scenario, the label represents an imitated sentiment tag. Before generating the next response, the appropriate sentiment tag is predicted to direct the generating process.

Our framework is expressive and extendable. The generated responses agree with the predefined labels while maintaining meaningful. By changing the definition of the label, our framework can be easily applied to other specific areas.

\section{Models}
To provide a better dialog context, we build a hierarchical recurrent encoder-decoder with separated context models (SPHRED).  This section first introduces the concept of SPHRED, then explains the conditional variational framework and two application scenarios.

\subsection{SPHRED}
\label{sec:sphred}
\begin{figure}
\centering
\centerline{\includegraphics[width=8.5cm]{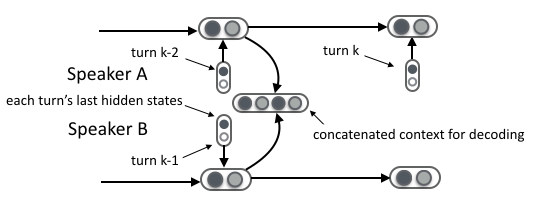}}
\caption{Computational graph for SPHRED structure. The status vector for Speaker A and Speaker B is modeled by separate RNNs then concatenated to represent the dialog context.}
\label{fig:sphred}
\end{figure}
We decomposes a dialog into two levels: sequences of utterances and sub-sequences of words, as in \cite{serban2015building}.
Let $\mathbf{w}_1, \ldots, \mathbf{w}_N$ be a dialog with $N$ utterances, where $\mathbf{w}_n = (w_{n,1}, \dots, w_{n,M_n})$ is the $n$-th utterance. The probability distribution of the utterance sequence factorizes as:
\begin{equation}
\prod_{n = 1}^N \prod_{m = 1}^{M_n} P_\theta(\mathbf{w}_{m,n} | \mathbf{w}_{m, < n}, \mathbf{w}_{< n})
\label{model: prob}
\end{equation}
where $\theta$ represents the model parameters and $\mathbf{w}_{< n}$ encodes the dialog context until step $n$.

If we model the dialog context through a single recurrent neural network (RNN), it can only represent a general dialog state in common but fail to capture the respective status for different speakers. This is inapplicable when we want to infer implicit personal attributes from it and use them to influence the sampling process of the latent variable, as we will see in Section \ref{sec: scene2}. Therefore, we model the dialog status for both speakers separately. As displayed in Figure \ref{fig:sphred}, SPHRED contains an encoder RNN of tokens and two status RNNs of utterances, each for one speaker. When modeling turn $k$ in a dialog, each status RNN takes as input the last encoder RNN state of turn $k-2$. The higher-level context vector is the concatenation of both status vectors.

We will show later that SPHRED not only well keeps individual features, but also provides a better holistic representation for the response decoder than normal HRED.

\subsection{Conditional Variational Framework}
\begin{figure}
\centering
\centerline{\includegraphics[width=4.2cm, height=2.7cm]{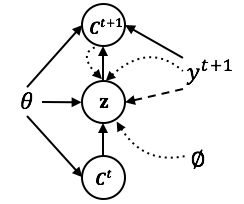}}
\caption{Graphical model for the conditional variational framework. Solid lines denote generative model $P_{\theta}(\mathbf{z}_n |\mathbf{y}_n,\mathbf{w}_1^{n-1})$ and $P_{\theta}(\mathbf{w}_n \mid \mathbf{y}_n, \mathbf{z}_n, \mathbf{w}_1^{n-1})$. When $y^{t+1}$  is known, there exists an additional link from $y^{t+1}$ to $z$ (dashed line). $C^t$ encodes context information up to time $t$. Dotted lines are posterior approximation $Q_{\phi}(\mathbf{z}_n|\mathbf{y}_n,\mathbf{w}_1^{n})$.}
\label{fig:generate}
\end{figure}

VAEs have been used for text generation in \cite{bowman2015generating,semeniuta2017hybrid}, where texts are synthesized from latent variables. Starting from this idea, we assume every utterance $\mathbf{w}_n$ comes with a corresponding label $\mathbf{y}_n$ and latent variable $\mathbf{z}_n$. The generation of $\mathbf{z}_n$ and $\mathbf{w}_n$ are conditioned on the dialog context provided by SPHRED, and this additional class label $\mathbf{y}_n$. This includes 2 situations, where the label of the next sequence is known (like for Scenario 1 in Section \ref{sec: scene1}) or not (Section \ref{sec: scene2}). For each utterance, the latent variable $\mathbf{z}_n$ is first sampled from a prior distribution. The whole dialog can be explained by the generative process:
\begin{equation}
P_{\theta}(\mathbf{z}_n|\mathbf{y_n},\mathbf{w}_1^{n-1}) = \mathcal{N}(\boldsymbol{\mu}_{\text{prior}}, \Sigma_{\text{prior}})
\end{equation}
\begin{equation}
\begin{aligned}
&P_{\theta}(\mathbf{w}_n \mid \mathbf{y}_n, \mathbf{z}_n, \mathbf{w}_1^{n-1}) = 
\\\prod_{m=1}^{M_n} P_{\theta}&(w_{n,m} \mid \mathbf{y}_n, \mathbf{z}_n,\mathbf{w}_1^{n-1}, w_{n,1}^{n,m-1})
\end{aligned}
\end{equation}
When the label $\mathbf{y}_n$ is unknown, a suitable classifier is implemented to first predict it from the context vector. This classifier can be designed as, but not restricted to, multilayer perceptrons (MLP) or support vector machines (SVM).

Similarly, the posterior distribution of $\mathbf{z}_n$ is approximated as in Equation \ref{formula: post}, where the context and label of the next utterance is provided. The graphical model is depicted in Figure \ref{fig:generate}.
\begin{equation}
\label{formula: post}
Q_{\phi}(\mathbf{z}_n|\mathbf{y}_n,\mathbf{w}_1^{n}) = \mathcal{N}(\boldsymbol{\mu}_{\text{posterior}}, \Sigma_{\text{posterior}})
\end{equation}

The training objective is derived as in Formula \ref{VHRED:lower_bound}, which is a lower bound of the logarithm of the sequence probability. When the label is to be predicted ($\mathbf{\bar{y}}_n$), an additional classification loss (first term) is added such that the distribution $q_{\phi}(\mathbf{y}_n|\mathbf{w}_1^{n-1})$ can be learned together with other parameters.
\begin{equation}
\begin{split}
&\log P_{\theta}(\mathbf{w}_1, \dots, \mathbf{w}_N)  \geq \mathbb{E}_{p(\mathbf{w}_n,\mathbf{y}_n)} \left[q_{\phi}(\mathbf{y}_n|\mathbf{w}_1^{n-1}) \right]
\\&- \sum_{n=1}^N \text{KL} \left [ Q_{\psi}(\mathbf{z}_n \mid \mathbf{w}_1^n,\mathbf{y}_n) || P_{\theta}(\mathbf{z}_n \mid\mathbf{w}_1^{n-1},{\mathbf{\bar{y}}_n}) \right ]
\\&+\mathbb{E}_{Q_{\psi}(\mathbf{z}_n\mid\mathbf{w}_1^n,\mathbf{y}_n)}[\log P_{\theta}(\mathbf{w}_n\mid\mathbf{z}_n,\mathbf{w}_1^{n-1},\mathbf{y}_n)] \label{VHRED:lower_bound}
\end{split}
\end{equation}

\subsection{Scenario 1}
\label{sec: scene1}
A major focus in the current research is to avoid generating generic responses, so in the first scenario, we let the label $\mathbf{y}$ indicate whether the corresponding sequence is a generic response, where $y = 1$ if the sequence is generic and $y = 0$ otherwise. To acquire these labels, we manually constructed a list of generic phrases like ``I have no idea", ``I don't know", etc. Sequences containing any one of such phrases are defined as generic, which in total constitute around 2 percent of the whole corpus. At test time, if the label is fixed as 0, we expect the generated response should mostly belong to the non-generic class.

No prediction is needed, thus the training cost does not contain the first item in Formula \ref{VHRED:lower_bound}. This scenario is designed to demonstrate our framework can explicitly control which class of responses to generate by assigning corresponding values to the label.

\subsection{Scenario 2}
\label{sec: scene2}
In the second scenario, we experiment with assigning imitated sentiment tags to generated responses. The personal sentiment is simulated by appending :), :( or :P at the end of each utterance, representing positive, negative or neutral sentiment respectively. For example, if we append ``:)'' to the original ``OK'', the resulting ``OK :)'' becomes positive. The initial utterance of every speaker is randomly tagged. We consider two rules for the tags of next utterances. Rule 1 confines the sentiment tag to stay constant for both speakers. Rule 2 assigns the sentiment tag of next utterance as the average of the preceding two ones. Namely, if one is positive and the other is negative, the next response would be neutral.

The label $\mathbf{y}$ represents the sentiment tag, which is unknown at test time and needs to be predicted from the context. The probability $q_{\phi}(\mathbf{y}_n|\mathbf{w}_1^{n-1})$ is modeled by feedforward neural networks. This scenario is designed to demonstrate our framework can successfully learn the manually defined rules to predict the proper label and decode responses conforming to this label.

\section{Experiments}
We conducted our experiments on the Ubuntu dialog Corpus~\cite{lowe2015ubuntu}, which contains about 500,000 multi-turn dialogs. The vocabulary was set as the most frequent 20,000 words. All the letters are transferred to lowercase and the Out-of-Vocabulary (OOV) words were preprocessed as $<$unk$>$ tokens.
\subsection{Training Procedures}
Model hyperparameters were set the same as in VHRED model except that we reduced by half the context RNN dimension. The encoder, context and decoder RNNs all make use of the Gated Recurrent Unit (GRU) structure~\cite{cho2014learning}. Labels were mapped to embeddings with size 100 and word vectors were initialized with the pubic Word2Vec embeddings trained on the Google News Corpus\footnote{\url{https://code.google.com/archive/p/word2vec/}}.  Following~\cite{bowman2015generating}, $25\%$ of the words in the decoder were randomly dropped. We multiplied the KL divergence and classification error by a scalar which starts from zero and gradually increases so that the training would initially focus on the stochastic latent variables. At test time, we outputted responses using beam search with beam size set to 5~\cite{graves2012sequence} and $<$unk$>$ tokens were prevented from being generated. We implemented all the models with the open-sourced Python library Tensorflow~\cite{abadi2016tensorflow} and optimized using the Adam optimizer~\cite{kingma2014adam}. Dialogs are cut into set of slices with each slice containing 80 words then fed into the GPU memory. All models were trained with batch size 128. We use the learning rate 0.0001 for our framework and 0.0002 for other models. Every model is tested on the validation dataset once every epoch and stops until it gains nothing more within 5 more epochs.

\subsection{Evaluation}

Accurate automatic evaluation of dialog generation is difficult~\cite{galley2015deltableu,pietquin2013survey}. In our experiment, we conducted three embedding-based evaluations (average, greedy and extrema)~\cite{liu2016not} on all our models, which map responses into vector space and compute the cosine similarity. Though not necessarily accurate, the embedding-based metrics can to a large extent measure the semantic similarity and test the ability of successfully generating a response sharing a similar topic with the golden answer. The results of a  GRU language model (LM), HRED and VHRED were also provided for comparison. For the two scenarios of our framework, we further measured the percentage of generated responses matching the correct labels (accuracy). In~\cite{liu2016not}, current popular metrics are shown to be not well correlated with human judgements. Therefore, we also carried out a human evaluation. 100 examples were randomly sampled from the test dataset. The generated responses from the models were shuffled and randomly distributed to 5 volunteers\footnote{All volunteers are well-educated students who have received a Bachelor's degree on computer science or above.}. People were requested to give a binary score to the response from 3 aspects, grammaticality, coherence with history context and diversity. Every response was evaluated 3 times and the result agreed by most people was adopted.

\subsection{Results of Metric-based Evaluation}
\label{sec: metric-results}
As can be seen from Table \ref{tab:metric scores}, SPHRED outperforms both HRED and LM over all the three embedding-based metrics. This implies separating the single-line context RNN into two independent parts can actually lead to a better context representation. It is worth mentioning the size of context RNN hidden states in SPHRED is only half of that in HRED, but it still behaves better with fewer parameters. Hence it is reasonable to apply this context information to our framework.

The last 4 rows in Table \ref{tab:metric scores} display the results of our framework applied in two scenarios mentioned in Section \ref{sec: scene1} and \ref{sec: scene2}. SCENE1-A and SCENE1-B correspond to Scenario 1 with the label fixed as 1 and 0. 90.9\% of generated responses in SCENE1-A are generic and 86.9\% in SCENE1-B are non-generic according to the manually-built rule, which verified the proper effect of the label. SCENE2-A and SCENE2-B correspond to rule 1 and 2 in Scenario 2. Both successfully predict the sentiment with very minor mismatches (0.2\% and 0.8\%). The high accuracy further demonstrated SPHRED's capability of maintaining individual context information. We also experimented by substituting the encoder with a normal HRED, the resulting model cannot predict the correct sentiment at all because the context information is highly mingled for both speakers. The embedding based scores of our framework are still comparable with SPHRED and even better than VHRED. Imposing an external label didn't bring any significant quality decline.

\begin{table}[!hbtp] \addtolength{\tabcolsep}{-2pt}  
    \footnotesize
    \centering
    \begin{tabular}{|l|c c c|c|}
        \hline
        \textbf{Model} & \textbf{Average} &\textbf{Greedy}& \textbf{Extrema} & \textbf{Accuracy} \\ \hline 
        LM & 0.360 & 0.348 & 0.310 & - \\ \hline
        HRED & 0.429 & 0.466 & 0.383 & - \\ \hline
        SPHRED & \textbf{0.468} & \textbf{0.478} & \textbf{0.434} & - \\ 			\hline\hline 
        VHRED & 0.403 & 0.432 & 0.374 & - \\ \hline
        SCENE1-A & - & - & - & 90.9\% \\ \hline
        SCENE1-B & 0.426 & 0.432 & 0.396 & 86.9\% \\ \hline
        SCENE2-A & 0.465 & 0.440 & 0.428 & 99.8\% \\ \hline
        SCENE2-B & 0.463 & 0.437 & 0.420 & 99.2\% \\ \hline
    \end{tabular}
    \caption{\label{tab:metric scores}Metric-based Evaluation. SCENE1-A is set to generate generic responses, so it makes no sense to measure it with embedding-based metrics}
\end{table}

\subsection{Results of Human Evaluation}

\begin{table*}
\centering
\begin{tabular}{p{95mm}|p{60mm}}
  \textbf{Context} & \textbf{Response}\\
  \hline
  { anybody in the house ?????\_\_eou\_\_ how to change the default ubuntu wall paper ? \_\_eou\_\_ how to change the default ubuntu wallpaper ? \_\_eou\_\_ \_\_eot\_\_ Is there an echo in your head ? Is there an echo in your head ? \_\_eou\_\_ \_\_eot\_\_  what do you mean ? \_\_eou\_\_ \_\_eot\_\_ Repeating = Bad . \_\_eou\_\_ \_\_eot\_\_ no body is answering me \_\_eou\_\_ \_\_eot\_\_} & { \color{white}.} \newline \textbf{LM:} \ What do you want to do with it ? \newline \textbf{HRED:} \ I don't know . \newline \textbf{SPHRED:} \ If you want to change the default wallpaper , you can change the default theme   \\ \hline
   {How can I install seamonkey ? \_\_eou\_\_ To save me the pastebin \_\_eou\_\_  I am looking to install seamonkey , anyone ? \_\_eou\_\_ \_\_eot\_\_ http://www.seamonkey-project.org/ \_\_eou\_\_ \_\_eot\_\_ It i snot in the ubuntu repository any more \_\_eou\_\_ \_\_eot\_\_} & { \color{white}.} \newline \textbf{SCENE1-A:} \ sorry i have no idea . \newline \textbf{SCENE1-B:} \ you need to find the package that you can use .\\ \hline
   {hey guys , how can I add an extra xsession to ubuntu 10.04 ? :) \_\_eou\_\_  that is , I dont want GNOME  :) \_\_eou\_\_ \_\_eot\_\_ try this : https://wiki.ubuntu. com/CustomXSession :( \_\_eou\_\_ \_\_eot\_\_} & { \color{white}.} \newline \textbf{SCENE2-A:} \ ok thanks :)   \\ \hline
   {hey guys , how can I add an extra xsession to ubuntu 10.04 ? :( \_\_eou\_\_  that is , I dont want GNOME  :( \_\_eou\_\_ \_\_eot\_\_ try this : https://wiki.ubuntu. com/CustomXSession :) \_\_eou\_\_ \_\_eot\_\_} & { \color{white}.} \newline \textbf{SCENE2-B:} \ thank you for the help ! :P   \\ \hline
\end{tabular}
\caption{{\label{tab: examples}}Examples of context-response pairs for the neural network models. \_\_eou\_\_ denotes end-of-utterance and \_\_eot\_\_ denotes end-of-turn token
  }
\end{table*}

We conducted human evaluations on VHRED and our framework (Table \ref{tab:human scores}). All models share similar scores, except SCENE1-A receiving lower scores with respect to coherence. This can be explained by the fact that SCENE1-A is trained to generate only generic responses, which limits its power of taking coherence into account. VHRED and Scenario 2 perform close to each other. Scenario 1, due to the effect of the label, receives extreme scores for diversity.

\begin{table}[!hbtp] \addtolength{\tabcolsep}{-2pt}
    \footnotesize
    \centering
    \begin{tabular}{|l|c|c|c|c|c|}
        \hline
        \textbf{Model} & \textbf{G} &\textbf{CD}& \textbf{C$\neg$D}&\textbf{$\neg$CD} &\textbf{$\neg$C$\neg$D}\\ \hline
        VHRED & 97\% & 41\% & 12 \% &24\%&23\%\\ \hline
        SCENE1-A & 96\% & 3\% &  37\% &1\%&59\%\\ \hline
        SCENE1-B & 96\% & 47\% &  9\% &40\%&4\%\\ \hline
        SCENE2-A & 97\% & 40\% & 14 \% &23\%&23\%\\ \hline
        SCENE2-B & 95\% & 38\% &  20\% &31\%&11\%\\ \hline
    \end{tabular}
    \caption{\label{tab:human scores}Human Judgements, G refers to Grammaticality and the last four columns is the confusion matrix with respect to coherence and diversity}
\end{table}

In general, the statistical results of human evaluations on sentence quality are very similar between the VHRED model and our framework. This agrees with the metric-based results and supports the conclusion drawn in Section \ref{sec: metric-results}. Though the sample size is relatively small and human judgements can be inevitably disturbed by subjective factors, we believe these results can shed some light on the understanding of our framework.

A snippet of the generated responses can be seen in Table \ref{tab: examples}. Generally speaking, SPHRED better captures the intentions of both speakers, while HRED updates the common context state and the main topic might gradually vanish for the different talking styles of speakers. SCENE1-A and SCENE1-B are designed to reply to a given context in two different ways. We can see both responses are reasonable and fit into the right class. The third and fourth rows are the same context with different appended sentiment tags and rules, both generate a suitable response and append the correct tag at the end.

\section{Discussion and future work}

In this work, we propose a conditional variational framework for dialog generation and verify it on two scenarios. To model the dialog state for both speakers separately, we first devised the SPHRED structure to provide the context vector for our framework. Our evaluation results show that SPHRED can itself provide a better context representation than HRED and help generate higher-quality responses. In both scenarios, our framework can successfully learn to generate responses in accordance with the predefined labels. Though with the restriction of an external label, the score of generated responses didn't significantly decreased, meaning that we can constrain the generation within a specific class while still maintaining the quality.

The manually-defined rules, though primitive, represent two most common sentiment shift conditions in reality. The results demonstrated the potential of our model. To apply to real-world scenarios, we only need to adapt the classifier to detect more complex sentiments, which we leave for future research. External models can be used for detecting generic responses or classifying sentiment categories instead of rule or symbol-based approximations. We focused on the controlling ability of our framework, future research can also  experiment with bringing external knowledge to improve the overall quality of generated responses.

\section{Acknowledgement}
This work was supported by the National Natural Science of China under Grant No. 61602451, 61672445 and JSPS KAKENHI Grant Numbers 15H02754, 16K12546.
\bibliography{acl2017}

\begin{thebibliography}{}
\expandafter\ifx\csname natexlab\endcsname\relax\def\natexlab#1{#1}\fi

\bibitem[{Abadi et~al.(2016)Abadi, Agarwal, Barham, Brevdo, Chen, Citro,
  Corrado, Davis, Dean, Devin et~al.}]{abadi2016tensorflow}
Mart{\'\i}n Abadi, Ashish Agarwal, Paul Barham, Eugene Brevdo, Zhifeng Chen,
  Craig Citro, Greg~S Corrado, Andy Davis, Jeffrey Dean, Matthieu Devin, et~al.
  2016.
\newblock Tensorflow: Large-scale machine learning on heterogeneous distributed
  systems.
\newblock {\em arXiv preprint arXiv:1603.04467\/} .

\bibitem[{Bowman et~al.(2015)Bowman, Vilnis, Vinyals, Dai, Jozefowicz, and
  Bengio}]{bowman2015generating}
Samuel~R Bowman, Luke Vilnis, Oriol Vinyals, Andrew~M Dai, Rafal Jozefowicz,
  and Samy Bengio. 2015.
\newblock Generating sentences from a continuous space.
\newblock {\em arXiv preprint arXiv:1511.06349\/} .

\bibitem[{Cho et~al.(2014)Cho, Van~Merri{\"e}nboer, Gulcehre, Bahdanau,
  Bougares, Schwenk, and Bengio}]{cho2014learning}
Kyunghyun Cho, Bart Van~Merri{\"e}nboer, Caglar Gulcehre, Dzmitry Bahdanau,
  Fethi Bougares, Holger Schwenk, and Yoshua Bengio. 2014.
\newblock Learning phrase representations using rnn encoder-decoder for
  statistical machine translation.
\newblock {\em arXiv preprint arXiv:1406.1078\/} .

\bibitem[{Galley et~al.(2015)Galley, Brockett, Sordoni, Ji, Auli, Quirk,
  Mitchell, Gao, and Dolan}]{galley2015deltableu}
Michel Galley, Chris Brockett, Alessandro Sordoni, Yangfeng Ji, Michael Auli,
  Chris Quirk, Margaret Mitchell, Jianfeng Gao, and Bill Dolan. 2015.
\newblock deltableu: A discriminative metric for generation tasks with
  intrinsically diverse targets.
\newblock {\em arXiv preprint arXiv:1506.06863\/} .

\bibitem[{Graves(2012)}]{graves2012sequence}
Alex Graves. 2012.
\newblock Sequence transduction with recurrent neural networks.
\newblock {\em arXiv preprint arXiv:1211.3711\/} .

\bibitem[{Kingma and Ba(2014)}]{kingma2014adam}
Diederik Kingma and Jimmy Ba. 2014.
\newblock Adam: A method for stochastic optimization.
\newblock {\em arXiv preprint arXiv:1412.6980\/} .

\bibitem[{Kingma et~al.(2014)Kingma, Mohamed, Rezende, and
  Welling}]{kingma2014semi}
Diederik~P Kingma, Shakir Mohamed, Danilo~Jimenez Rezende, and Max Welling.
  2014.
\newblock Semi-supervised learning with deep generative models.
\newblock In {\em Advances in Neural Information Processing Systems\/}. pages
  3581--3589.

\bibitem[{Kingma and Welling(2013)}]{kingma2013auto}
Diederik~P Kingma and Max Welling. 2013.
\newblock Auto-encoding variational bayes.
\newblock {\em arXiv preprint arXiv:1312.6114\/} .

\bibitem[{Liu et~al.(2016)Liu, Lowe, Serban, Noseworthy, Charlin, and
  Pineau}]{liu2016not}
Chia-Wei Liu, Ryan Lowe, Iulian~V Serban, Michael Noseworthy, Laurent Charlin,
  and Joelle Pineau. 2016.
\newblock How not to evaluate your dialogue system: An empirical study of
  unsupervised evaluation metrics for dialogue response generation.
\newblock {\em arXiv preprint arXiv:1603.08023\/} .

\bibitem[{Lowe et~al.(2015)Lowe, Pow, Serban, and Pineau}]{lowe2015ubuntu}
Ryan Lowe, Nissan Pow, Iulian Serban, and Joelle Pineau. 2015.
\newblock The ubuntu dialogue corpus: A large dataset for research in
  unstructured multi-turn dialogue systems.
\newblock {\em arXiv preprint arXiv:1506.08909\/} .

\bibitem[{Pietquin and Hastie(2013)}]{pietquin2013survey}
Olivier Pietquin and Helen Hastie. 2013.
\newblock A survey on metrics for the evaluation of user simulations.
\newblock {\em The knowledge engineering review\/} 28(01):59--73.

\bibitem[{Rezende et~al.(2014)Rezende, Mohamed, and
  Wierstra}]{rezende2014stochastic}
Danilo~Jimenez Rezende, Shakir Mohamed, and Daan Wierstra. 2014.
\newblock Stochastic backpropagation and approximate inference in deep
  generative models.
\newblock {\em arXiv preprint arXiv:1401.4082\/} .

\bibitem[{Semeniuta et~al.(2017)Semeniuta, Severyn, and
  Barth}]{semeniuta2017hybrid}
Stanislau Semeniuta, Aliaksei Severyn, and Erhardt Barth. 2017.
\newblock A hybrid convolutional variational autoencoder for text generation.
\newblock {\em arXiv preprint arXiv:1702.02390\/} .

\bibitem[{Serban et~al.(2016)Serban, Sordoni, Bengio, Courville, and
  Pineau}]{serban2015building}
Iulian~V Serban, Alessandro Sordoni, Yoshua Bengio, Aaron Courville, and Joelle
  Pineau. 2016.
\newblock Building end-to-end dialogue systems using generative hierarchical
  neural network models.
\newblock {\em AAAI\/} .

\bibitem[{Serban et~al.(2017)Serban, Sordoni, Lowe, Charlin, Pineau, Courville,
  and Bengio}]{serban2016hierarchical}
Iulian~Vlad Serban, Alessandro Sordoni, Ryan Lowe, Laurent Charlin, Joelle
  Pineau, Aaron Courville, and Yoshua Bengio. 2017.
\newblock A hierarchical latent variable encoder-decoder model for generating
  dialogues.
\newblock {\em AAAI\/} .

\bibitem[{Shang et~al.(2015)Shang, Lu, and Li}]{shang2015neural}
Lifeng Shang, Zhengdong Lu, and Hang Li. 2015.
\newblock Neural responding machine for short-text conversation.
\newblock {\em arXiv preprint arXiv:1503.02364\/} .

\bibitem[{Sohn et~al.(2015)Sohn, Lee, and Yan}]{sohn2015learning}
Kihyuk Sohn, Honglak Lee, and Xinchen Yan. 2015.
\newblock Learning structured output representation using deep conditional
  generative models.
\newblock In {\em Advances in Neural Information Processing Systems\/}. pages
  3483--3491.

\bibitem[{Sordoni et~al.(2015)Sordoni, Galley, Auli, Brockett, Ji, Mitchell,
  Nie, Gao, and Dolan}]{sordoni2015neural}
Alessandro Sordoni, Michel Galley, Michael Auli, Chris Brockett, Yangfeng Ji,
  Margaret Mitchell, Jian-Yun Nie, Jianfeng Gao, and Bill Dolan. 2015.
\newblock A neural network approach to context-sensitive generation of
  conversational responses.
\newblock {\em arXiv preprint arXiv:1506.06714\/} .

\bibitem[{Sutskever et~al.(2014)Sutskever, Vinyals, and
  Le}]{sutskever2014sequence}
Ilya Sutskever, Oriol Vinyals, and Quoc~V Le. 2014.
\newblock Sequence to sequence learning with neural networks.
\newblock In {\em Advances in neural information processing systems\/}. pages
  3104--3112.

\bibitem[{Vinyals and Le(2015)}]{vinyals2015neural}
Oriol Vinyals and Quoc Le. 2015.
\newblock A neural conversational model.
\newblock {\em arXiv preprint arXiv:1506.05869\/} .

\bibitem[{Yan et~al.(2016)Yan, Yang, Sohn, and Lee}]{yan2016attribute2image}
Xinchen Yan, Jimei Yang, Kihyuk Sohn, and Honglak Lee. 2016.
\newblock Attribute2image: Conditional image generation from visual attributes.
\newblock In {\em European Conference on Computer Vision\/}. Springer, pages
  776--791.

\bibitem[{Yao et~al.(2015)Yao, Zweig, and Peng}]{yao2015attention}
Kaisheng Yao, Geoffrey Zweig, and Baolin Peng. 2015.
\newblock Attention with intention for a neural network conversation model.
\newblock {\em arXiv preprint arXiv:1510.08565\/} .

\end{thebibliography}
\bibliographystyle{acl_natbib}

\appendix

\end{document}